\documentclass[a4paper,fleqn]{cas-dc}

\usepackage[numbers]{natbib}
\usepackage{subfig}
\usepackage[ruled,linesnumbered]{algorithm2e}
\def\tsc#1{\csdef{#1}{\textsc{\lowercase{#1}}\xspace}}
\tsc{WGM}
\tsc{QE}
\tsc{EP}
\tsc{PMS}
\tsc{BEC}
\tsc{DE}

\begin{document}
\let\WriteBookmarks\relax
\def\floatpagepagefraction{1}
\def\textpagefraction{.001}
\shortauthors{C. Hu et~al.}

\title [mode = title]{IFNSO: Iteration-Free Newton–Schulz Orthogonalization}                      

\tnotetext[1]{This work was supported by the National Natural Science Foundation of China(Grant No,U21B2090).}
\tnotetext[2]{The paper is under consideration at Pattern Recognition Letters}
\author[1]{Chen Hu}[orcid=0009-0006-0876-5026]
\ead{huch67@mail2.sysu.edu.cn}
\credit{Writing – review \& editing, Writing – original draft, Visualization, Methodology, Conceptualization}
\author[1]{Qianxi Zhao}[orcid=0000-0002-8247-0713]
\credit{Data curation, Writing – review \& editing}
\ead{zhaoqx9@mail2.sysu.edu.cn}
\author[2]{Xiaochen Yuan}
\credit{Visualization, Investigation}
\ead{xcyuan@mpu.edu.mo}
\author[3]{Hong Zhang}
\credit{Visualization, Investigation}
\ead{dmrzhang@buaa.edu.cn}
\author[3]{Ding Yuan}
\credit{Formal analysis}
\ead{dyuan@buaa.edu.cn}
\author[1]{Yanbin Wu}
\credit{Writing – review \& editing}
\ead{wuyb53@mail2.sysu.edu.cn}
\author[1]{Xiying Li}[orcid=0000-0002-4753-8022]
\cormark[1]
\cortext[1]{Corresponding author}
\credit{Supervision, Project administration, Funding acquisition}
\ead{stslxy@mail.sysu.edu.cn}


\affiliation[1]{organization={School of Intelligent Systems Engineering},
            addressline={Sun Yat-sen University}, 
            city={Shenzhen},
            postcode={518107}, 
            state={Guangdong},
            country={PR China}}
\affiliation[2]{organization={Faculty of Applied Sciences},
            addressline={Macao Polytechnic University}, 
            city={Macao},
            postcode={999078}, 
            state={Macao},
            country={PR China}}
\affiliation[3]{organization={School of Astronautics},
            addressline={Beihang
University}, 
            city={Beijing},
            postcode={100191}, 
            state={Beijing},
            country={PR China}}

\begin{abstract}
The Newton–Schulz (NS) iteration has become a key technique for orthogonalization in optimizers such as Muon and for optimization on the Stiefel manifold. 
Despite its effectiveness, the conventional NS iteration incurs significant computational overhead due to repeated high-dimensional matrix multiplications. 
To overcome these limitations, we propose Iteration-Free Newton–Schulz Orthogonalization (IFNSO), a novel framework that replaces the traditional iterative structure with a unified and Iteration-Free structure with learnable coefficients. 
In order to optimize these coefficients stably, we introduce a new polynomial which is constrained to 1 as the input approaches 1.
Additionally, we employ an exponential term selection strategy, an analysis of the contribution of each term, to further improve performance.
Extensive experiments demonstrate that IFNSO achieves superior performance compared to existing methods. Our code is available at: \url{https://github.com/greekinRoma/Ieration_Free_Newton_Schulz_Orthogonalization}.
\end{abstract}

\begin{highlights}
\item IFNSO: A unified operation that reduces matrix multiplication burden.
\item A novel polynomial ensures stable convergence when optimizing coefficients.
\item Term exponents scale exponentially for more effective orthogonalization.
\end{highlights}

\begin{keywords}
Newton–Schulz (NS) iteration\sep
Iteration-Free Newton–Schulz Orthogonalization (IFNSO)\sep
Unified framework\sep 
Singular value decomposition\sep
\end{keywords}

\maketitle

\section{Introduction}
Orthogonalization has been widely adopted across diverse domains, including neural network optimization \cite{khaled2025muonbpfastermuonblockperiodic, bernstein2024modular, pethick2025training, he2025root, ai2025practicalefficiencymuonpretraining, boreiko2025towards} and Riemannian optimization on the Stiefel manifold \cite{tunccel2009optimization, chen2023decentralized, gao2021riemannian}.
In particular, the Muon optimizer serves as a representative orthogonalization-based optimizer, demonstrating improved performance compared with Adam \cite{kingma2017adammethodstochasticoptimization} in large-scale language model pretraining \cite{page2025muonallmuonvariantefficient}, and is increasingly adopted in large-scale model training.

\begin{figure}
    \centering
    \subfloat[Original NS iteration.]{\includegraphics[width=0.2\textwidth]{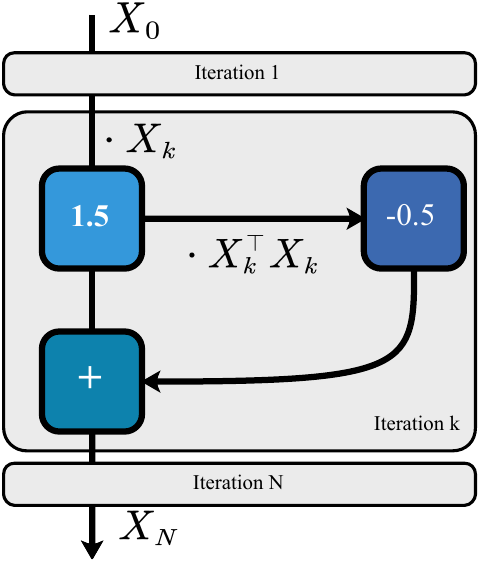}\label{fig:method-a}}
    \subfloat[Muon's NS iteration.]{\includegraphics[width=0.2\textwidth]{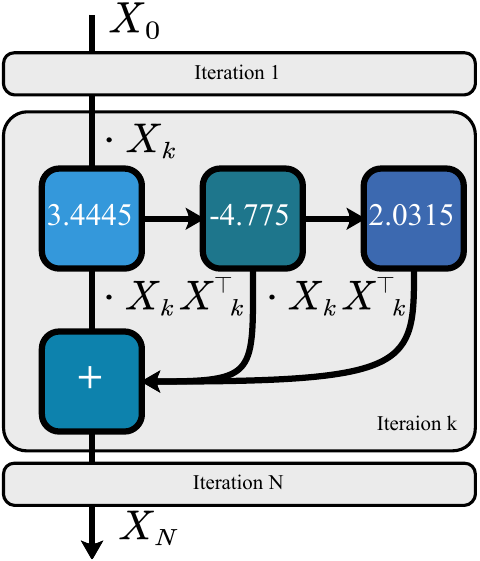}\label{fig:method-muon}}\\
    \subfloat[Cesista's NS iteration.]{\includegraphics[width=0.2\textwidth]{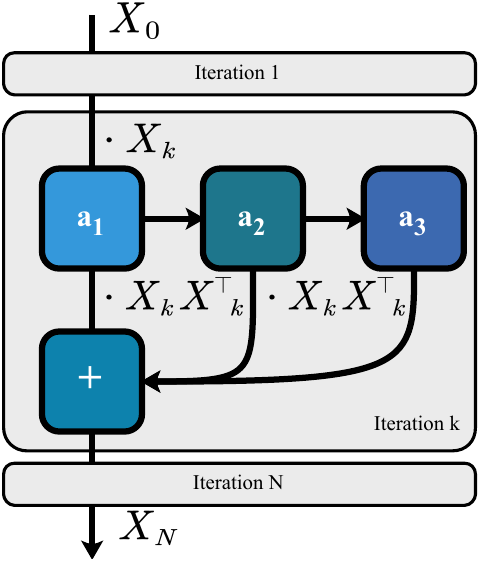}\label{fig:method-c}}
    \subfloat[CANS.]{\includegraphics[width=0.2\textwidth]{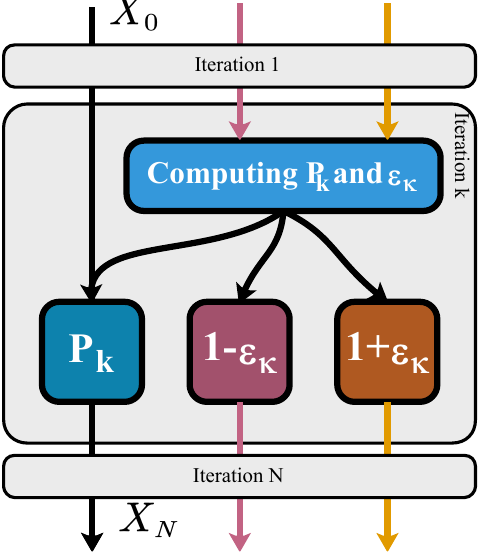}\label{fig:method-d}}
    \caption{The existing NS iteration methods, where $\epsilon_k$ is hyperparameter.}
    \label{fig:ns-methods}
\end{figure}

One of the key techniques for orthogonalization is the Newton–Schulz (NS) iteration
\cite{bjorck1971iterative, kovarik1970some}, which replaces singular value decomposition (SVD) with matrix multiplications, thereby reducing computational cost. 
Figure \ref{fig:method-a} illustrates the procedure of the classical NS iteration \cite{higham2008functions}.

First, the input matrix $M \in \mathbb{R}^{h_{\text{input}} \times w_{\text{input}}}$ is transposed to ensure that the height is less than or equal to the width. We define $A \in \mathbb{R}^{h \times w}$ as:
\begin{equation}
    A = 
    \begin{cases} 
        M & \text{if } h_{\text{input}} \leq w_{\text{input}} \\
        M^\top & \text{if } h_{\text{input}} > w_{\text{input}}.
    \end{cases}
\end{equation}
This transformation ensures $h \le w$, thereby minimizing the size of the product $AA^\top \in \mathbb{R}^{h \times h}$.

To guarantee convergence of the NS iteration, the singular values of the initial matrix must satisfy $\sigma(X_0) < 1$. We initialize the iteration input $X_0$ by normalizing $A$ using its Frobenius norm \cite{nakatsukasa2012backward}:
\begin{equation}
    X_0 = \frac{A}{\|A\|_F}.
    \label{constraint}
\end{equation}
where $\|\cdot\|_F$ denotes the Frobenius norm.

Finally, the method iteratively refines the matrix through a sequence of updates. For the iteration index $k = 0, 1, \dots, N-1$, the Newton–Schulz iteration is defined as:
\begin{equation}
    X_{k+1} = \frac{1}{2}  \left( 3I - X_k X_k^\top \right) X_k,
    \label{one_iteration_matrix}
\end{equation}
where $I \in \mathbb{R}^{h \times h}$ denotes the identity matrix. Specifically, if the Singular Value Decomposition (SVD) of $X_0$ is given by $U S_0 V^\top$, where $U\in \mathbb{R}^{h\times h}$ and $V \in \mathbb{R}^{w\times h}$ are orthonormal matrices.

Since $U$ and $V$ are unchanged during the iteration, each singular value $\sigma$ evolves independently via the scalar map:
\begin{equation}
    g(\sigma) = 1.5 \sigma -0.5 \sigma^3.
    \label{one_iteration_scalar}
\end{equation}
By recursively applying the refinement operator defined in Eq. \eqref{one_iteration_scalar}, the sequence $g^N(\sigma)$ converges toward $\operatorname{sign}(\sigma)$ for normalized $\sigma$. Specifically, in the limit $N \to \infty$, the $N$-fold composition of $g$ approximates the following function:
\begin{equation}
    \lim_{N \to \infty} g^N(\sigma) = \underbrace{g(g(\cdots g(\sigma)\cdots))}_{N \text{ times}} \rightarrow 1.
    \label{N_fold}
\end{equation}

By iteratively applying this operator, the processed representation $X_N$ converges toward $UV^\top$. Specifically, let $S = \operatorname{diag}(\sigma_1, \dots, \sigma_h)$ represent the singular values. Then $X_N$ is derived as follows:
\begin{equation}
\begin{aligned}
    X_N &= U g^N(S) V^{\top} \\
        &= U \operatorname{diag}\left(g^N(\sigma_1), \dots, g^N(\sigma_h)\right) V^{\top} \\
        &\xrightarrow{N \to \infty} U I V^{\top} \\
        &= U V^{\top}.
\end{aligned}
\end{equation}

More generally, each iteration can be written as:
\begin{equation}
X_{k+1} = \sum_{n=0}^{M} a_{2n+1} (X_k X_k^\top)^n X_k,
\label{general_eq_matrix}
\end{equation}
where $a_{n}$ is the coefficient and $M$ determines the maximum degree.
Like the relation between Eq. \eqref{one_iteration_matrix} and Eq. \eqref{one_iteration_scalar}, the corresponding scalar formula is shown as below:
\begin{equation}
x_{k+1}=p(x_k) = \sum_{n=0}^{M} a_{2n+1} x_k^{2n+1}.
\label{general_eq_scalar}
\end{equation}

Building upon Eq. \eqref{general_eq_matrix} and Eq. \eqref{general_eq_scalar}, a number of studies have been conducted to improve its performance. 
Higham and Schreiber \cite{higham1990fast} proposed an adaptive hybrid iteration that accelerates polar decomposition on modern hardware.

As shown in Figure \ref{fig:method-muon}, Keller Jordan \cite{jordan2024muon} sets $M$ to 2 and proposes a new form of NS iteration based on previous works \cite{bernstein2024old, guo2006schur}:
\begin{equation}
    X_{k+1} = a_0 X_k + a_3 (X_k X_k^\top) X_k + a_5 (X_k X_k^\top)^2 X_k.
    \label{iteration_jordan}
\end{equation}
Additionally, Keller Jordan proposed standard parameters to achieve better convergence: ($a_0$, $a_3$, $a_5$) = (3.4445, -4.7750, 2.0315) for Muon optimizer.

Other studies \cite{cesista2025muonoptcoeffs} have incorporated iteration-specific learnable coefficients. These coefficients are derived by optimizing a reparameterized scalar function, defined as follows:
\begin{align}
x_{k+1} &= \gamma \cdot \prod_{i=0}^{4} (x_k - f_i) + x_k, \\
f &= [-(1 + r), -(1 - u), 0, 1 - u, 1 + r],
\end{align}
where $\gamma$, $r$, and $u$ are learnable parameters that are optimized to improve the efficiency of the polynomial. These parameters are then used to derive the coefficients $a_0$, $a_3$, and $a_5$ in Figure \ref{fig:method-c}. 

Grishina et al. \cite{grishina2025accelerating} propose a new polynomial based on the Chebyshev alternation theorem, called Chebyshev-accelerated Newton–Schulz (CANS), as shown in Figure \ref{fig:method-d}. They also use Gelfand's formula:
\begin{equation}
\sigma_1(A) \leqslant \|(A A^\top)^q\|_F^{1/(2q)}, \label{gelfand_formula}
\end{equation}
where \(\sigma_1(A)\) is the largest singular value of \(A\) and $q$ is the hyperparameter determined by researchers. Based on this bound, we scale the input matrix as follows:
\begin{equation}
X_0 = \frac{A}{\|(A A^\top)^q\|_F^{1/(2q)}},
\label{scale_formula_function}
\end{equation}
which ensures that the singular values of \(X_0\) fall within the range (0,1), as given by:
\begin{equation}
\sigma(X_0) = \frac{\sigma_i(A)}{\sigma_1(A)} \leqslant 1.
\end{equation}

Although the above algorithms have been proposed to improve orthogonalization efficiency, they still rely on iterative structures, hindering the further improvements in efficiency due to the repeated matrix multiplication along the long dimension. 
We introduce a non-iterative structure, named Iteration-Free Newton–Schulz Orthogonalization (IFNSO), to improve efficiency by replacing iterative steps with a single unified operation \cite{stotsky2019unified}. 
In the unified formula, we sum each term with learnable coefficients, whose solution space is large.
In order to optimize the coefficients stably, we propose a new polynomial with a constraint.
Additionally, we analyze the role of each matrix with varied exponents, discard negligible terms, and use learnable coefficients to get the optimized polynomial. 
Based on a series of experiments, our approach consistently outperforms existing methods.
To summarize, the contributions of this manuscript are outlined below.
\begin{itemize}
    \item [1.] We propose the IFNSO, a single-pass operator with learnable parameters designed to replace traditional iterative blocks, reducing the number of matrix multiplications along the long dimension from N to 1.
    \item[2.] We adopt a new polynomial with a constraint that ensures its curve approaches 1 when the input tends to 1, providing a more stable gradient flow, which accelerates the convergence of the learnable coefficients.
    \item[3.] We analyze the contribution of terms with varied exponents and introduce an exponential growth strategy for our polynomial to achieve a better performance in efficiency.
\end{itemize}

\section{Method}
We scale our matrix according to Eq. \eqref{scale_formula_function}, where we set $q=1$ for simplicity:
\begin{equation}
    X_0=\frac{A}{\|(A A^\top)\|_F^{1/2}}.
    \label{new_scale_formula}
\end{equation}
The proposed IFNSO algorithm is then applied to orthogonalize the matrix, and the details of IFNSO are described in the following sections.
\begin{figure}
    \centering
    \subfloat[Linear Growth\label{fig:subfig-b}]{
        \includegraphics[width=0.95\linewidth]{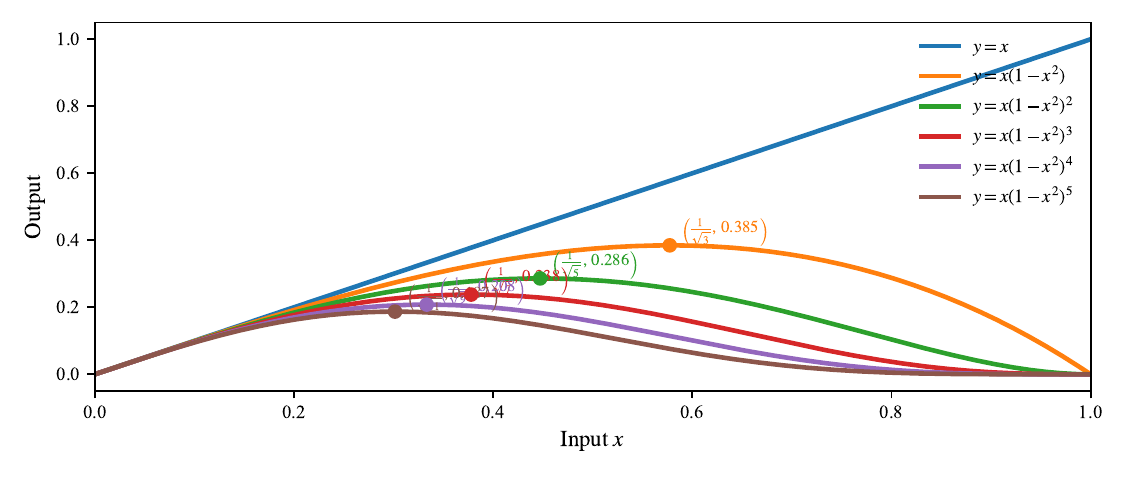}
    }\\
     \subfloat[Exponential Growth\label{fig:subfig-a}]{
        \includegraphics[width=0.95\linewidth]{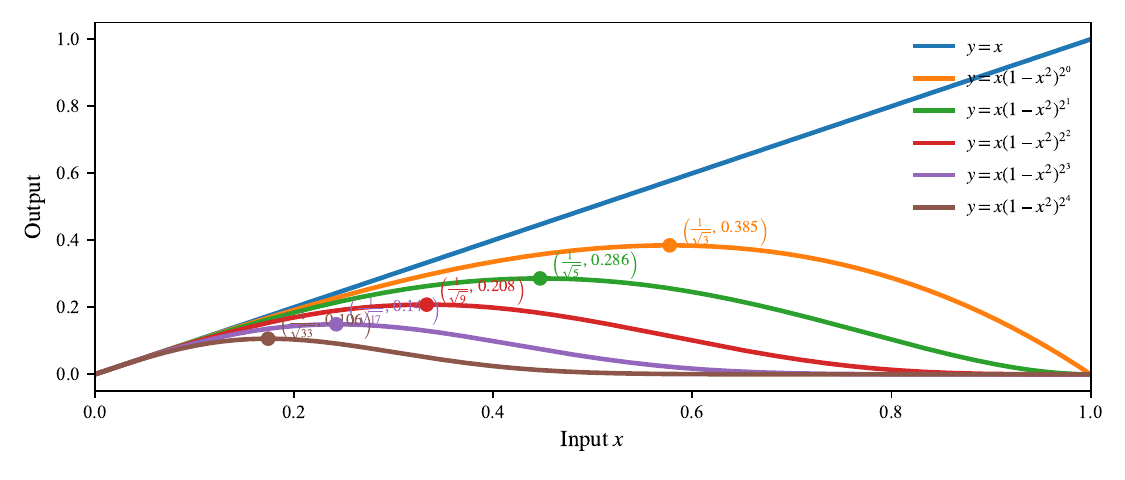}
    }
    \caption{Term selection under linear and exponential growth of $n_l$.}
    \label{fig:paraneter_pattern}
\end{figure}

\subsection{Iteration-Free  Polynomial}
By recursively substituting the iterative updates, all intermediate terms can be expressed as a single polynomial in $x_0$. 
The original Newton–Schulz (NS) iteration can be represented according to Eq.~\eqref{N_fold} and Eq. \eqref{general_eq_scalar}:
\begin{equation}
    x_N = \sum_{n=0}^{\frac{(2M+1)^N-1}{2}} b_{2n+1} x_0^{2n+1}.
    \label{polynomial_unify}
\end{equation}
The matrix form of Eq. \eqref{polynomial_unify} is:
\begin{equation}
X_N = \sum_{n=0}^{\frac{(2M+1)^N-1}{2}} b_{2n+1} (X_0 X_0^\top)^n X_0,
\label{old_unified_formula}
\end{equation}
where \( b_{2n+1} \) are coefficients determined by \( a_n \) in Eq.~\eqref{general_eq_matrix}. 

To simplify the expression and eliminate the iterative structure, we replace \( X_N \) with \( Y \), \( X_0 \) with \( X \), $\frac{(2M+1)^N-1}{2}$ with $K$ and $b_i$ with learnable $w_i$, and obtain the following formula:
\begin{equation}
    Y = \sum_{n=0}^{K} w_{2n+1} (X X^\top)^n X.
\label{new_unified_formula_0}
\end{equation}

The corresponding scalar formula of Eq. \eqref{new_unified_formula_0} is as shown below:
\begin{equation}
    y = \sum_{n=0}^{K} w_{2n+1} x^{2n+1}.
    \label{new_unified_formula}
\end{equation}
where $w_{2n+1}$ are obtained through optimization.
To ensure stable convergence during optimization, we impose constraints on the coefficients:
\begin{equation}
    y = \sum_{n=0}^{K} w_{2n+1} (1-x^2)^n\times x,
    \label{our_scalar_formula}
\end{equation}
where the coefficients $w_{2n+1}$ are constrained such that $y \to 1$ as $x \to 1$.

The corresponding matrix equation of Eq. \eqref{our_scalar_formula} is as follows:
\begin{equation}
Y =  \sum^K_{n=0}w_{2n+1}(I-XX^\top)^nX.
\label{after_matrix_formula}
\end{equation}

In practice, the expansion terms in Eq.~\eqref{after_matrix_formula} are not strictly required to follow a consecutive sequence of exponents. 
To allow non-consecutive or sparse selection of $n$, we introduce a discrete index set ${n_0, n_1, \dots, n_L}$ to generalize the selection:
\begin{equation}
    Y = \sum_{l=0}^{L} w_l(I - XX^\top)^{n_l} X.
    \label{generalized_matrix_formula}
\end{equation}

Each term in Eq. \eqref{generalized_matrix_formula} corresponds to repeatedly applying the projection operator $(I - X X^{\top})$ to $X$, yielding a polynomial structure.
For the singular values, the contribution of the $l$-th term can be expressed as
\begin{equation}
y =f_{l}(x)= x(1 - x^2)^{n_l}.
\label{final_polynomial}
\end{equation}

\subsection{Term Selection}
According to Eq. \eqref{final_polynomial}, the sequence ${n_l}$ is sparse.
We first investigate the selection of $\{n_l\}_{l=0}^L$, which governs the performance and computational efficiency. 

In the following sections, we explore several scheduling strategies for $n_l$, including uniform linear increments and exponential growth, to study the trade-off between approximation accuracy and inference latency.

As shown in Figure \ref{fig:paraneter_pattern}, exponential growth expands the range of powers much faster than linear growth, enabling broader coverage with the same number of terms.
Based on this observation, we adopt an exponential growth in parameterization, defined as
\begin{equation}
n_l =
\begin{cases}
0, & l = 0, \\
2^{l-1}, & l = 1, \dots, L .
\end{cases}
\end{equation}
After that, Eq.~\eqref{final_polynomial} can be rewritten as
\begin{equation}
y = f_l(x) =
\begin{cases}
x, & l = 0, \\
x\left(1 - x^2\right)^{2^{l-1}}, & l= 1, \dots, L
\end{cases}
\label{final_polynomial_term}
\end{equation}
We compute the gradient of Eq. \eqref{final_polynomial_term}.
When $l=0$, the gradient of $f_0(x)$ is $1$.
when $l>0$,  the gradient is given by Eq. \eqref{eq_gradient}.
\begin{equation}
\begin{aligned}
f_l'(x) = \left(1 - x^2\right)^{2^{l-1}-1} \left[ 1 - (2^l + 1)x^2 \right].
\end{aligned}
\label{eq_gradient}
\end{equation}

For \(l > 0\), the function \(f_l(x)\) has an extreme point within the interval \((0,1)\), given by the expression
\begin{equation}
\begin{aligned}
    x^* &= \frac{1}{\sqrt{2^l+1}}, \\
    y^* &= \frac{1}{\sqrt{2^l+1}}(\frac{2^l}{2^l + 1})^{2^{l-1}}\approx \frac{1}{\sqrt{2^l+1}}e^{-\frac{1}{2}}.
\end{aligned}
\end{equation}
As $l$ increases, the extreme point moves closer to $0$, indicating that the polynomial forms a sharper and more localized peak, as shown in Figure \ref{fig:k_increase_pattern}. 
\subsection{Coefficient Optimization}
\begin{figure}
    \subfloat[\label{fig:subfig-a}]{
        \includegraphics[width=0.45\linewidth]{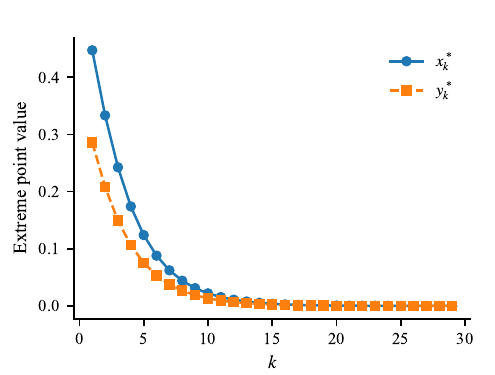}
    }
    \subfloat[\label{fig:subfig-b}]{
        \includegraphics[width=0.45\linewidth]{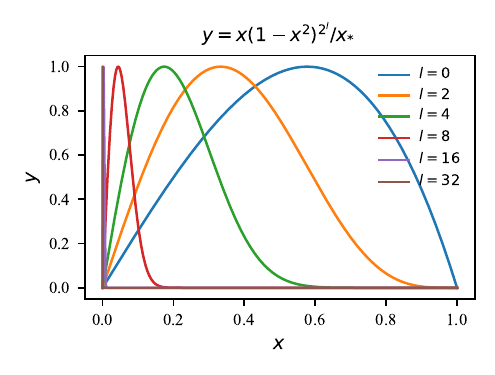}
    }
    \caption{Family of curves for increasing $l$. (a) Extreme-point coordinates $(x^*, y^*)$; (b) Normalized response $y/y^*$.}
    \label{fig:k_increase_pattern}
\end{figure}

According to the above analysis, our formula becomes
\begin{equation}
\begin{aligned}
     f(x)= x + \sum_{l=1}^Lw_lx\left(1 - x^2\right)^{2^{l-1}}.
     \label{sum_equation}
\end{aligned}
\end{equation}
As illustrated in Figure~\ref{fig:k_increase_pattern}, the $L$-th term dominates the convergence toward $1$. 
Hence, we set the $x$-coordinate of the extreme point of $f_L(\cdot)$ as the location where the Eq. \eqref{sum_equation} first reaches $1$, and obtain the following constraints:
\begin{equation}
    f(\frac{1}{\sqrt{2^L+1}}) = 1
    \label{peak}
\end{equation}
According to Eq. \eqref{sum_equation} and \eqref{peak}, $w_L$ could be represented as follows:
\begin{equation}
\begin{aligned}
w_L &= \left( \sqrt{2^L+1} - 1 
      - \sum_{k=1}^{L-1} w_k \left( \frac{2^L}{2^L+1} \right)^{2^{k-1}} \right) \\
    &\quad \times \left( \frac{2^L+1}{2^L} \right)^{2^{L-1}} .
\end{aligned}
\end{equation}
The above expression is complex, and its exact expression can be approximated by a simpler form:
\begin{equation}
    w_L \approx e^{1/2} \left( 2^{L/2} - 1 \right) - \sum_{l=1}^{L-1} w_l.
    \label{b_approximation}
\end{equation}

The coefficients $w_l$ $(l=1,\dots,L-1)$ are learnable through optimization.

\subsection{Overall Structure}
By leveraging the property $(I - XX^\top)^{2^{l-1}} = ((I - XX^\top)^{2^{l-2}})^2$, we can compute higher-order terms using fewer matrix multiplications.

The polynomial form of IFNSO used during training to obtain the learnable parameters $w_l$ $(l=1,\dots,L-1)$ is given by
\begin{equation}
\begin{aligned}
f(x) &= x + \sum_{l=1}^{L-1} w_l\, x (1 - x^2)^{2^{l-1}} \\
     &\quad + \left(e^{1/2}\!\left(2^{L/2}-1\right) - \sum_{l=1}^{L-1} w_l\right)
        x(1-x^2)^{2^{L-1}} .
\end{aligned}
\label{overall_polynomial}
\end{equation}

\begin{figure}
    \centering
    \includegraphics[width=0.8\linewidth]{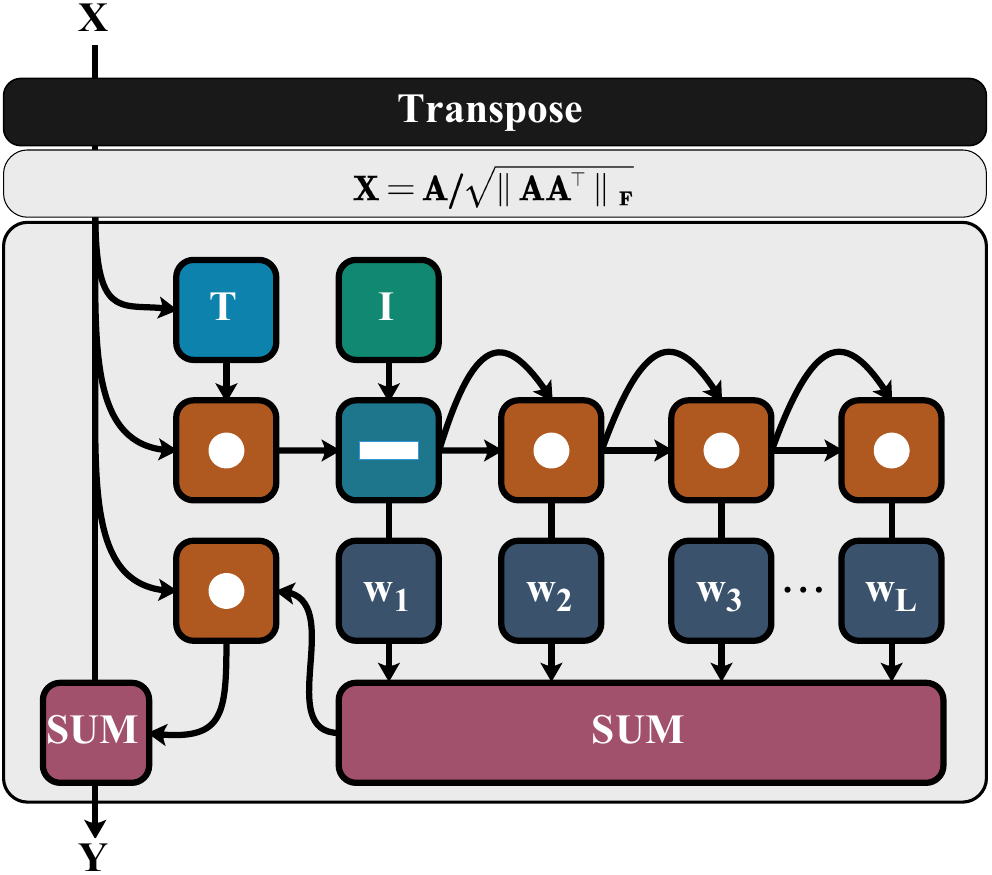}
    \caption{The overall structure of IFNSO.}
    \label{fig:overall_structure}
\end{figure}

All coefficients in $f(\cdot)$ are learned by minimizing the following objective:
\begin{equation}
\min_{w_1,\dots,w_{L-1}}
\sum_{j=1}^{J} \left(f(x_j)-1\right)^2,
\quad \text{where } x_j \in (0,1).
\end{equation}
Here, $\{x_j\}_{j=1}^{J}$ denotes $J$ sampled points from the interval $(0,1)$.
Once the coefficients are obtained, the input matrix can then be orthogonalized accordingly, as illustrated in Figure~\ref{fig:overall_structure}.
The corresponding matrix formula is 
\begin{equation}
   Y = X + \sum_{l=1}^{L} w_l\, (1 - XX^\top)^{2^{l-1}}X,\quad X=\frac{A}{\| AA^\top\|^{\frac{1}{2}}_F}.
   \label{newnew_formula}
\end{equation}

To efficiently obtain $Y$, we combine Eq.~\eqref{new_scale_formula} with $XX^\top$ and define a sequence of matrices $\{T_l\}_{l=0}^{L}$ as
\begin{equation}
    T_l=
    \begin{cases}
    AA^\top,&l=0,\\
	I- T_0/\| T_0 \|_F, & l=1, \\
	T_{l-1}T_{l-1}, & l>1.
    \end{cases}
\end{equation}
Then, we aggregate all $T_l$ to obtain $Y$ by transforming Eq. \eqref{newnew_formula}:
\begin{align}
Y' &= I + \sum^{L}_{l=1}w_lT_l,\\
Y  &= \frac{Y' A}{\sqrt{\| T_0 \|_F}},
\end{align}
where $Y'$ is an intermediate matrix.

To sum up, the above formula is implemented by explicitly constructing each polynomial term through matrix multiplications, as summarized in Algorithm~\ref{alg:poly_iteration}.
\begin{algorithm}[h]
\caption{IFNSO (Proposed Method)}
\label{alg:poly_iteration}

\KwIn{
Matrix $M \in \mathbb{R}^{h_{\text{input}} \times w_{\text{input}}}$; \\
Learned coefficients $\{w_l\}_{l=1}^L$;\\
Identity matrix $I$.
}

\KwOut{
Output matrix $Y$
}

\textbf{Coefficient learning:}

$\displaystyle 
\min_{\{w_l\}_{l=1}^{L-1}}
\sum_{j=1}^{J} \left(f(x_j)-1\right)^2,
\quad \text{where } f(\cdot) \text{ is defined in Eq.~\eqref{overall_polynomial}}.
$

\textbf{Initialization:}

$A \leftarrow 
\begin{cases}
M^\top, & h_{\text{input}} > w_{\text{input}},\\
M, & \text{otherwise}
\end{cases}$

$T_0 = AA^\top$\\
$o=\| T_0 \|_F$ \\
$T_1 \leftarrow I - T_0/o$

\textbf{Polynomial expansion:}

\For{$l = 2$ \KwTo $L$}{
    $T_l \leftarrow T_{l-1}\cdot T_{l-1}$
}

\textbf{Polynomial aggregation:}

$Y' \leftarrow I + \sum_{l=1}^{L} w_l T_l$

$Y \leftarrow Y' A/\sqrt{o}$

\Return{$Y$}

\end{algorithm}

\section{Experiment}
\subsection{Experimental Setup}
All experiments are implemented in Python 3.10 with PyTorch 2.1 and executed on a single NVIDIA RTX 4050 GPU.
Experiments are performed on a Windows system with CUDA~11.8 for GPU acceleration.

\subsubsection{Optimization of $\{ w_l\}_{l=1}^{L-1}$}
We employ the Adam optimizer with an initial learning rate of $10^{-1}$. The learning rate is decayed by a factor of $0.5$ every $10000$ iterations using a step scheduler. Training is conducted for $20000$ epochs. To approximate the expectation at each iteration, the loss is computed by uniformly sampling $1000$ points from the interval $(0, 1)$.

\subsection{Experimental Setup for Orthogonalizing the Matrices}
We conduct experiments to evaluate the orthogonalization performance on matrices of different sizes. Specifically, we randomly generate matrices with dimensions $(128\times128)$, $(128\times512)$, and $(128\times1024)$.
The evaluation metrics include Floating-Point Operations (FLOPs) and Error defined as
\begin{equation}
\mathrm{\textbf{Error}} = \sqrt{\sum_{i=1}^{h}\sum_{j=1}^{h} E_{i,j}^{2}},
\quad \text{where} \quad
E = Y Y^\top - I .
\end{equation}

\subsubsection{Experimental Setup for Application Testing}
\label{application_testing}
We conduct a comparative analysis between our proposed algorithm and other NS iteration methods to demonstrate the efficacy of our approach. 
The network is trained on the MNIST dataset \cite{lecun2010mnist} using the Muon optimizer coupled with varied NS iteration algorithms. 
We adopt a 6:1 ratio for the training and testing split. 
For the training parameters, we select cross-entropy loss as the objective function, setting the batch size to 32 and capping the maximum number of epochs at 10.

\subsection{Ablation Study}
\begin{figure}
    \centering
     \includegraphics[width=\linewidth]{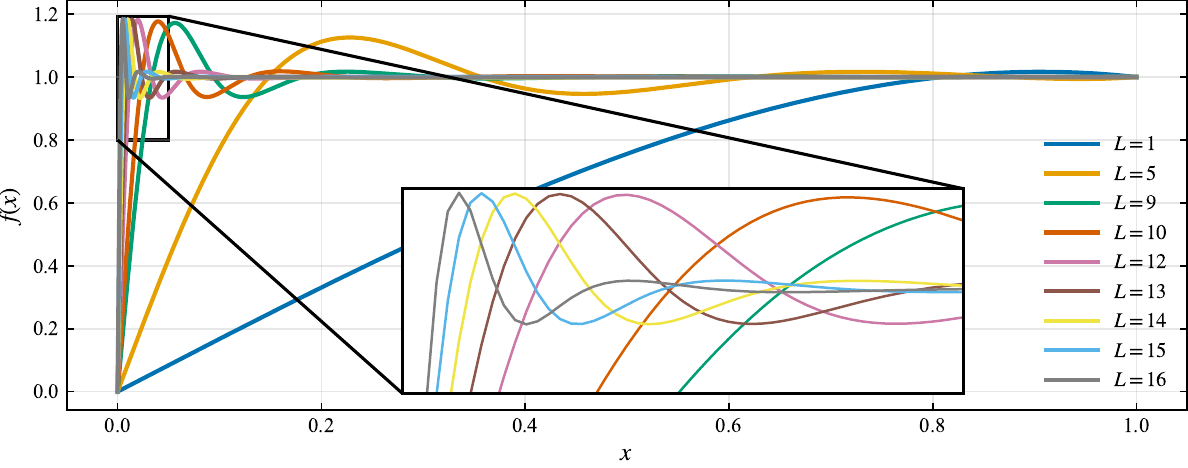}
    \caption{The optimized curve with the different $L$.}
    \label{fig:Ablation Study}
\end{figure}

We study the effect of the polynomial depth $L$ on approximation performance, as shown in Figure \ref{fig:Ablation Study}.
By varying $L$ while keeping all other training settings fixed, we observe that increasing $L$ consistently improves the expressive capacity of the polynomial, leading to lower approximation error on the target interval $(0,1)$.
However, beyond a certain depth, the performance gain saturates, while the computational burden gradually increases.
Therefore, we set $14$ as our default $L$ for the matrix.
These results indicate that a moderate value of $L$ provides the best trade-off between approximation accuracy and stability.

\begin{figure}
    \centering
    \includegraphics[width=\linewidth]{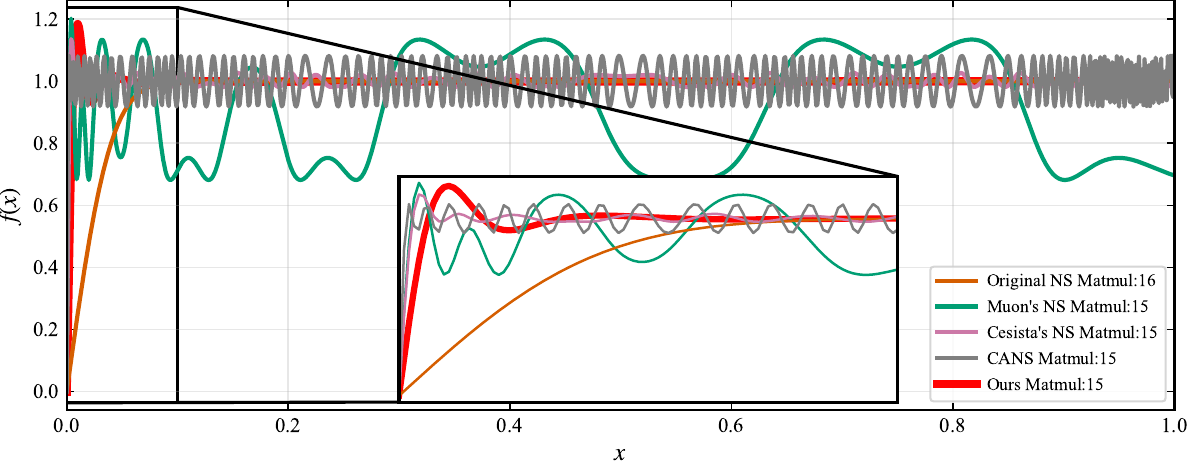}
    \caption{The curves based on different algorithms.}
    \label{fig:Comparative_experiment}
\end{figure}

\subsection{Comparative Experiments}
We compare our proposed method against Origin NS, Muon’s NS\cite{jordan2024muon}, Cesista’s NS\cite{cesista2025muonoptcoeffs}, and the CANS algorithm\cite{grishina2025accelerating}.
\subsubsection{Comparative Experiments Based on Performance Curves}
We perform a curve-based comparative evaluation to assess both the functional behavior and practical computational cost of different curve construction methods. As shown in Figure \ref{fig:Comparative_experiment}, our curve is compared with the original curves of other methods over the same input domain. While our method reaches 1 at x=0.005, slight fluctuations are observed near the convergence point.

\begin{table}
\centering
\caption{Performance of the different orthogonalization methods with increasing matrix size.}
\begin{tabular}{c c c c}
\toprule
\textbf{Method} &
\textbf{Size ($h\times w$)} &
\textbf{Error $\downarrow$} &
\textbf{FLOPs $\downarrow$} \\
\midrule
\multirow{3}{*}{Original NS} 
&$128 \times 128$  & $0.4869$ & $6.750\times10^{7}$ \\
&$128 \times 512$  & $0.4884$ & $2.696\times10^{8}$ \\
&$128 \times 1024$ & $0.4970$ & $5.391\times10^{8}$ \\
\hline
\multirow{3}{*}{Muon's NS} 
&$128 \times 128$  & $3.846$ & $6.332\times10^{7}$ \\
&$128 \times 512$  & $3.838$ & $2.533\times10^{8}$ \\
&$128 \times 1024$ & $3.844$ & $5.066\times10^{8}$ \\
\hline
\multirow{3}{*}{Cesista's NS} 
&$128 \times 128$  & $0.330$ & $6.332\times10^{7}$ \\
&$128 \times 512$  & $0.330$ & $2.533\times10^{8}$ \\
&$128 \times 1024$ & $0.330$ & $5.066\times10^{8}$ \\
\hline
\multirow{3}{*}{CANS} 
&$128 \times 128$  & $1.311$ & $6.332\times10^{7}$ \\
&$128 \times 512$  & $1.309$ & $2.533\times10^{8}$ \\
&$128 \times 1024$ & $1.311$ & $5.066\times10^{8}$ \\
\hline
\multirow{3}{*}{Our} 
&$128 \times 128$  & $0.040$ & $6.314\times10^{7}$ \\
&$128 \times 512$  & $0.040$ & $8.831\times10^{7}$ \\
&$128 \times 1024$ & $0.040$ & $1.219\times10^{8}$ \\
\bottomrule
\end{tabular}
\label{table_1}
\end{table}
\begin{table}
\centering
\caption{Performance of different NS iteration strategies applied to training a simple neural network on the MNIST dataset. We bold the best result.}
\begin{tabular}{c c c}  
\toprule
\textbf{Method} &
\textbf{Loss ($\times10^{-2}$)$\downarrow$} &
\textbf{Accuracy (\%) $\uparrow$} \\  
\midrule
Original NS &6.78&98.52\\
Muon's NS &5.10 &98.83 \\
Cesista's NS &5.47 & 98.78\\
CANS & 10.62 & 97.97\\
Our &\textbf{4.25} &\textbf{98.87}\\
\bottomrule
\end{tabular}
\label{table_2}
\end{table}

\subsubsection{Comparative Experiments on Practical Matrix Operations}
As illustrated in Tab.~\ref{table_1}, our method demonstrates competitive efficiency by reducing the number of computationally expensive $XX^\top$ operations from $N$ to $1$, where $XX^\top$ is more costly than $(XX^\top)(XX^\top)$ because $h \le w$.
This advantage stems from its lower cost of floating-point computations, achieved by reducing the amount of computation along the $w$-axis.
Specifically, as $w$ increases, the computational burden of our method grows more slowly compared to other methods.
Our method also achieves competitive orthogonalization performance.
\subsubsection{Comparative Experiments on Optimization Performance}
\begin{figure}
    \centering
    \includegraphics[width=\linewidth]{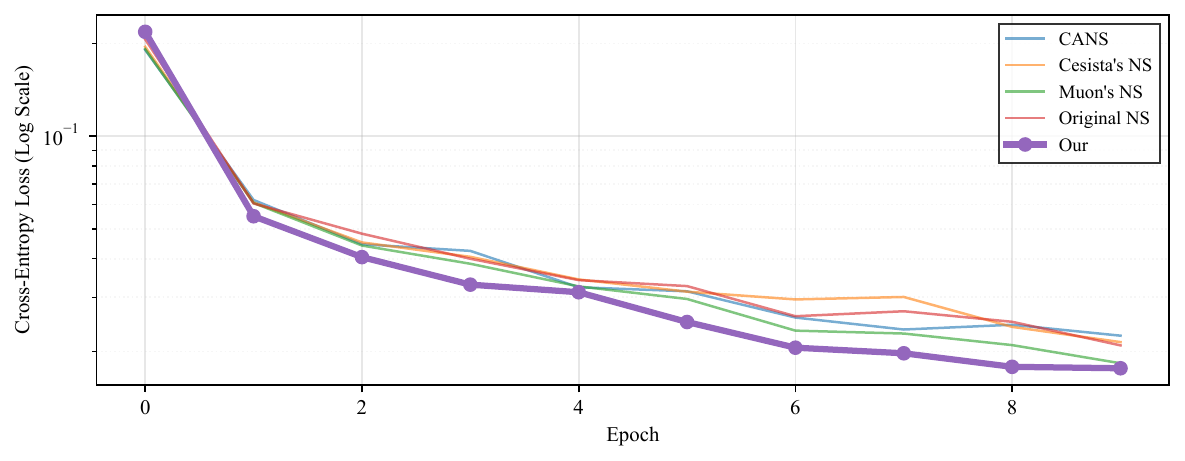}
    \caption{Comparison of training loss on the MNIST dataset using the Muon optimizer with various NS iteration strategies.}
    \label{fig:training}
\end{figure}

Following the experimental setup detailed in Section \ref{application_testing}, we compare our proposed strategy (IFSNO) with existing methods applied to the Muon optimizer for training a simple neural network on the MNIST dataset. As shown in Table \ref{table_2}, the network optimized using our strategy yields the best performance. Correspondingly, Figure \ref{fig:training} demonstrates that the model trained with our NS iteration algorithm achieves the fastest convergence.

\section{Conclusion}
We propose IFNSO, a method that reduces repeated computation along the longer dimension ($w$).
Experiments indicate that the proposed method achieves competitive performance with lower computational cost.
However, several limitations remain to be addressed. 
First, our method converges to $1$ at a slower rate, compared with some methods. 
Second, noticeable fluctuations appear when $y$ first reaches $1$.
Finally, the matrix multiplications involving $X$ are still required, leading to high computation cost when $w\gg h$.
In the future, we aim to overcome these limitations by introducing dynamic coefficients to enable a more flexible term structure.
Besides, we plan to explore neural network-based techniques to reduce the effective dimension of $w$ via low-rank approximations.

\printcredits

\bibliographystyle{cas-model2-names}

\bibliography{cas-refs}

\end{document}